\documentclass[review]{elsarticle}
\usepackage{svg}
\usepackage{amsmath}
\usepackage{lineno,hyperref}
\modulolinenumbers[1]

\usepackage[absolute,overlay]{textpos}

\journal{Cognition}

\begin{document}

\begin{frontmatter}

\begin{textblock*}{20cm}(3.5cm,3cm) 
\sffamily To appear in Cognition. This preprint may contain errors not present in
the published version.
\end{textblock*}

\title{Evolution of emotion semantics}

\author[a]{Aotao Xu\corref{mycorrespondingauthor}}\cortext[mycorrespondingauthor]{Corresponding author}
\ead{a26xu@cs.toronto.edu}
\author[b]{Jennifer E. Stellar} 
\author[a,c]{Yang Xu}

\address[a]{Department of Computer Science, University of Toronto}
\address[b]{Department of Psychology, University of Toronto}
\address[c]{Cognitive Science Program, University of Toronto}

\begin{abstract}
Humans possess the unique ability to communicate emotions through language. Although concepts like anger or awe are abstract, there is a shared consensus about what these English emotion words mean. This consensus may give the impression that their meaning is static, but we propose this is not the case. We cannot travel back to earlier periods to study emotion concepts directly, but we can examine text corpora, which have partially preserved the meaning of emotion words. Using natural language processing of historical text, we found evidence for semantic change in emotion words over the past century and that varying rates of change were predicted in part by an emotion concept's prototypicality---how representative it is of the broader category of ``emotion''. Prototypicality negatively correlated with historical rates of emotion semantic change obtained from text-based word embeddings, beyond more established variables including usage frequency in English and a second comparison language, French. This effect for prototypicality did not consistently extend to the semantic category of birds, suggesting its relevance for predicting semantic change may be category-dependent. Our results suggest emotion semantics are evolving over time, with prototypical emotion words remaining semantically stable, while other emotion words evolve more freely.

\end{abstract}


\end{frontmatter}

\linenumbers

\section{Introduction}

Much like emotion {concepts} vary in their meaning across cultures~\cite{kitayama1994emotion,wierzbicka1999emotions,jackson2019emotion}, it is possible emotion words can take on different meanings over time.\footnote{In our study, we use the terms ``emotion concept'' and ``emotion word'' interchangeably to refer to emotions that are lexicalized in natural language.} For instance, the English word {\it awe} in the 18th century may not represent the same feeling it does today, after a century of evolving perspectives on power and beauty~\cite{keltner2003approach}. Although we cannot travel to earlier historical periods to study emotion concepts directly, we do have access to text corpora which have partially preserved the meaning of emotion words. These words do not reflect the entirety of an emotion concept, which includes expressive, experiential, and physiological components, but they do offer insight into its shared meaning within a society. Here we use computational linguistic analyses to investigate the evolution of emotion semantics.

\begin{figure}[ht]
\centering
\includesvg[scale=0.5]{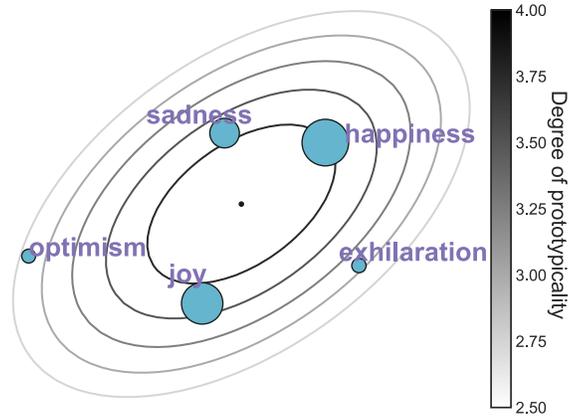}
\caption{An illustration of  the relation between prototypicality and semantic stability of emotion words. Each blue dot represents an emotion word, and the size of the dot is proportional to its predicted semantic stability; the smaller the dot, the higher its rate of semantic change over time. The contours indicate degrees of prototypicality. Visually, an emotion word close to the center has high prototypicality, and vice versa.
} \label{fig:illustration}
\end{figure}

If the meaning of different emotion words like \textit{awe} or \textit{joy} are changing over time, are they changing at the same rate or are there features of an emotion word that might predict its rate of semantic change? We propose that an emotion's conceptual prototypicality is one such feature. Prototypicality is a graded measure of the goodness of a concept's membership in a semantic category~\cite{fehr1984concept,shaver1987emotion}. In the case of emotions, \textit{joy} is considered a more prototypical concept than \textit{optimism}. Prototypical emotion concepts may have clearer biological and cultural functions and more distinct features than less prototypical ones. For instance, prototypical concepts like {\it fear} and {\it disgust} are particularly suited to solving evolutionary challenges or taking advantage of opportunities that faced early humans~\cite{cosmides2000evolutionary}, and they may have particularly strong social or cultural scripts~\cite{barrett2006emotions,fehr1984concept,russell1999core}. These emotion concepts are often more clearly marked by distinctive expressions, experience, and patterns of activation in the body~\cite{ekman2011basic}, and prototypical members may even help define the meaning of their less prototypical counterparts~\cite{johnson1989language} (see {\it Supplementary Information} for further evidence). We hypothesize that these well-defined functions and features of prototypical emotion concepts could promote semantic stability. As a result, the meaning of words for more prototypical concepts like {\it joy} may tend to resist change, more so than words for less prototypical ones like \textit{optimism}; see Figure~\ref{fig:illustration} for an illustration. 


Although prototypicality has been discussed in other semantic categories, we do not expect  prototypicality to predict semantic stability in every category. The basis of prototypicality and thus its ability to predict semantic change may differ in the classic example of birds~\cite{rosch1975cognitive}. The prototypicality of a bird name is primarily based on differences in biological taxonomies~\cite{boster1988natural} and features grounded in sensory or visual perception~\cite{garrard2001prototypicality}. As such, the features that define more (e.g., \textit{sparrow}) or less prototypical birds (e.g., \textit{penguin}) are equally well-defined, so the meanings of prototypical bird names do not help define the meanings of less prototypical bird names (see {\it Supplementary Information}), in contrast to the category of emotion words. We expect that while prototypicality should correlate with semantic stability of emotion words, it should not correlate with semantic stability of bird names. 


Our hypothesis augments general principles of lexical evolution and semantic change. It has been shown that usage frequency is a general determiner of stability in English verb regularization~\cite{lieberman2007quantifying}, lexical replacement~\cite{pagel2007frequency,pagel2013ultraconserved,vejdemo2016semantic}, loan word borrowing~\cite{monaghan2019cognitive}, and semantic change~\cite{hamilton2016diachronic,dubossarsky2017outta}. If explained through the lens of communication, frequency should predict semantic stability: when speakers change the meaning of a highly frequent lexical item, they would face a higher number of misunderstandings than if they change a low-frequency item~\cite{boyd1988culture,bybee2007frequency}. As a result, we expect frequent emotion words to change less in meaning than other emotion words. We examine the prototypicality of emotion concepts as an additional predictor of semantic stability beyond usage frequency. 

Our hypothesis differs from diachronic prototype semantics~\cite{geeraerts1997diachronic}, which states that more prototypical senses of a word tend to stay prototypical over time and exhibit more stability than peripheral senses. Although this theory is consistent with our hypothesis regarding the pattern that prototypicality offers stability, we focus on explaining rates of semantic change among concepts in a lexical field, as opposed to characterizing principles of change among  senses of an individual word~\cite{geeraerts1997diachronic,ramiro2018algorithms}. Previous studies have examined the theory of diachronic prototype semantics over the whole lexicon and found the prototypicality of words in statistical clusters (formed in meaning space) to negatively correlate with rates of semantic change~\cite{dubossarsky2015bottom,dubossarsky2017outta}. However, these studies do not explain how semantic change relates to prototypicality in the scope of a specific category such as emotions or birds. 

We present a methodology for modeling emotion semantics and its evolution by building on work from machine learning and natural language processing in word embedding~\cite{mikolovA,mikolov2013distributed,caliskan2017semantics} and its historical extensions~\cite{kulkarni2015statistically,hamilton2016diachronic,xie19,li2019macroscope}. We model emotion semantics using a vector-space representation trained on historical text corpora of natural language use, and we use this representation to model human judgments of prototypicality and semantic change of emotion words. Vector-space models of word meaning have been used within affective science for reconstructing human emotion ratings on dimensions such as valence and arousal~\cite{buechel2018word}, sentiment analysis~\cite{mohammad2016sentiment}, and analyzing emotion categories in documents~\cite{calvo2013emotions}, but not for investigating the open question of the evolution of emotion semantics.


\section{Methodologies for quantifying rates of semantic change}

Quantifying the rate of semantic change for a word requires records of its meaning from two distinct time periods and a quantitative metric that compares these records. One type of methods that constructs word meanings and enables comparisons over time is based on word embeddings~\cite{mikolovA,mikolov2013distributed}. The embedding of a word is a real-valued vector that represents its meaning through a high-dimensional space; vectors for words with similar meanings tend to be close in this space, such as {\it compassion} and {\it sympathy}. Word embeddings are constructed from co-occurrence statistics in large text corpora. We thus obtain meaning representations from two distinct time periods by constructing word embeddings based on historical text corpora from the corresponding periods~\cite{hamilton2016diachronic}. 

Existing methods for computing rates of semantic change often rely on the cosine distance between two embeddings~\cite{hamilton2016diachronic, dubossarsky2017outta}. According to this metric, a large cosine distance between historical and contemporary embeddings implies a high rate of semantic change, and vice versa. However, this metric tends to bias the correlation between rate of semantic change and frequency~\cite{dubossarsky2017outta}. For this reason, we use an alternate neighbourhood-based metric to compare word embeddings across time~\cite{xu2015computational}. This metric quantifies the rate of semantic change for a word $w$ between periods $t_1$ and $t_2$ via the Jaccard distance between sets of $k$-nearest neighbours in meaning space:
\begin{equation}
    rate(w, t_1, t_2) = 1 - \frac{| kNN(w, t_1) \, \cap \, kNN(w, t_2) |}{| kNN(w, t_1) \, \cup \, kNN(w, t_2) |} \label{eq:1}
\end{equation}
\noindent where $kNN(w,t)$ contains the $k$ words whose embeddings are the closest to the embedding of $w$ in terms of cosine similarity. Intuitively, we say a word underwent semantic change if the composition of its semantic neighbourhood has changed. Following~\cite{xu2015computational}, the part of speech (POS) of the members of $kNN(w,t)$ is always the same as the POS of $w$, and we also set $k$ to 100. In {\it Supplementary Information}, we show that this measure is robust to variations in $k$. Compared to the cosine metric, this metric enables more transparent interpretation on rates of change because we can inspect and evaluate the sets of semantic neighbours (see {\it Supplementary Information} for examples of emotion semantic change).


To implement this metric at scale, we used pretrained historical word embeddings and POS tags from HistWords~\cite{hamilton2016diachronic}. Specifically, we used 300-dimensional Word2Vec (SGNS) embeddings obtained from the Skip-Gram model~\cite{mikolov2013distributed} and trained on the corpora Google Books Ngrams English and French. We used historically most frequent POS tags from the same sources. This provided us with historical word embeddings and most frequent POS tags for 100,000 English words and 100,000 French words, for every decade between 1800 and 2000. 


\section{Analyses of emotion concepts}
In the first set of analyses, we provide evidence for our hypothesis  that the well-defined features and functions of prototypical emotion words promote semantic stability. Specifically, we test against the null hypothesis that prototypicality does not predict semantic stability in English and French emotion words over the past century.\footnote{We focused on these two languages because 1) we want to test if our analysis generalizes beyond a single language, and 2) there is a limited cross-linguistic variety of empirical studies on emotion prototypicality and of the historical data provided by HistWords.} We describe resources that provide us with lists of English and French emotion words, emotion prototypicality ratings, and historical frequency estimates. We then describe our methods for estimating prototypicality ratings historically and for hypothesis testing, which is followed by a presentation of our results.

\subsection{Materials}
We obtained a list of English emotion words and their corresponding prototypicality ratings from~\cite{shaver1987emotion}. The authors produced the list by obtaining 213 emotion nouns from a collection of emotion concepts. They produced emotion prototypicality ratings by asking 112 American university students to rate each of these nouns on a 4-point scale, where 4 means the noun is definitely an emotion, and 1 means the noun is definitely not an emotion. Following this work, our analyses focused on nouns that have prototypicality ratings at least 2.75 with the addition of {\it surprise} and exclusion of {\it abhorrence}, {\it ire}, {\it malevolence}, and {\it titillation}; we additionally included the word {\it awe}. We also obtained the valence of these emotion words from the study, which was originally derived from applying multidimensional scaling to similarity judgments~\cite{shaver1987emotion}.

We also obtained a list of French emotion words with their corresponding prototypicality ratings~\cite{niedenthal2004prototype}. The authors produced the list by translating 237 Italian emotion words from an earlier study into French. They produced emotion prototypicality ratings by asking 319 French university students to rate each of these words on a 10-point scale, where 10 means the word is certainly an emotion, and 1 means it is not an emotion. To be consistent with the English list, we kept emotion words whose most frequent POS tag is noun in the final decade of our historical POS data. We also obtained the valence of these emotion words from the study, which was originally obtained by asking 300 French university students to rate the words on a scale of -5 (very unpleasant) to 5 (very pleasant)~\cite{niedenthal2004prototype}.

We obtained historical frequency data from HistWords~\cite{hamilton2016diachronic}, which is based on the corpora Google Books Ngrams English and French. This yielded historical frequency data for 682,459 English words and 213,686 French words, for every decade between 1800 and 2000. We intersected the word lists with historical word embeddings, POS tags, and frequency from HistWords. We noticed that more emotion words were unavailable when we increased the span between flanking decades than otherwise: if $t_1 = 1890$ and $t_2 = 1990$, only 9 words from the English list are unavailable in HistWords and the HTE, but if we used $t_1 = 1800$, the number increased to 28; similarly in French, the shorter time span resulted in 32 unavailable words, but the longer one resulted in 58 unavailable words. Consequently, we decided to use the decades of 1890 and 1990 as the flanking decades for our analysis (i.e. $t_1 = 1890$, $t_2 = 1990$), and we used historical frequency data from the 1890s. After the intersection, we had a total of 123 English emotion words and 112 French emotion words.

\subsection{Methods}
Since we cannot go back in time to measure the prototypicality of emotion concepts in the past, we needed a method for estimating historical prototypicality. Let $x$ represent the word embedding of a concept in category $c$. Following previous work in prototype theory~\cite{reed1972pattern,ashby1995categorization}, we estimated the prototypicality of $x$ as the unnormalized conditional probability $p(c | x)$, which can be computed using an isotropic Gaussian via Bayes rule:
\begin{equation}
    p(c | x) \propto p(x | c) \sim N(\mu, I) \label{eq:2}
\end{equation}
where $\mu = \frac{1}{|E_c|}\sum_{v \in E_c} v$ and $E_c$ is the set of embeddings for members of $c$; $I$ is an identity matrix. Intuitively, we estimated the prototypicality of $x$ by computing its distance from the category centroid $\mu$; the closer they are, the higher its estimated prototypicality. To estimate the prototypicality of an emotion concept in history, we used its historical embedding $x$ and the embeddings of other emotion concepts to compute $p(x | c = emotion)$. We evaluated this method by computing the correlation between our empirical prototypicality ratings obtained from~\cite{shaver1987emotion,niedenthal2004prototype} and our estimated prototypicality based on embeddings from the 1980s and 1990s, the decades closest to the publication of those studies.

To test against the null hypothesis, we computed the rate of change for every emotion concept $x$, $rate(x, 1890, 1990)$ using Equation~\eqref{eq:1} and historical embeddings and POS tags from HistWords. Separately for English and French, we then computed the Pearson correlations between the emotion concepts' rates of change and prototypicality estimated for the 1890s. To evaluate whether the prototypicality of emotion concepts predicts rates of change beyond frequency, we performed multiple linear regressions for English and French using the following regression formula: 
\begin{equation}
    rate(x, 1890, 1990) \sim p(x | c=emotion) + freq(x) + val(x) 
    \label{eq:3}
\end{equation}
where for every concept $x$, we denote its usage frequency as $freq(x)$ and its valence as $val(x)$, which we added to control for unequal numbers of negative and positive emotion concepts in our datasets. We fitted the model using ordinary least squares implemented by {\texttt {statsmodel}}~\cite{seabold2010statsmodels}; we also used this package to compute relevant test statistics. Following previous work~\cite{hamilton2016diachronic}, we performed a log transformation on frequency.

\subsection{Results}
\begin{figure}[ht]
\centering
\includesvg[width=\linewidth]{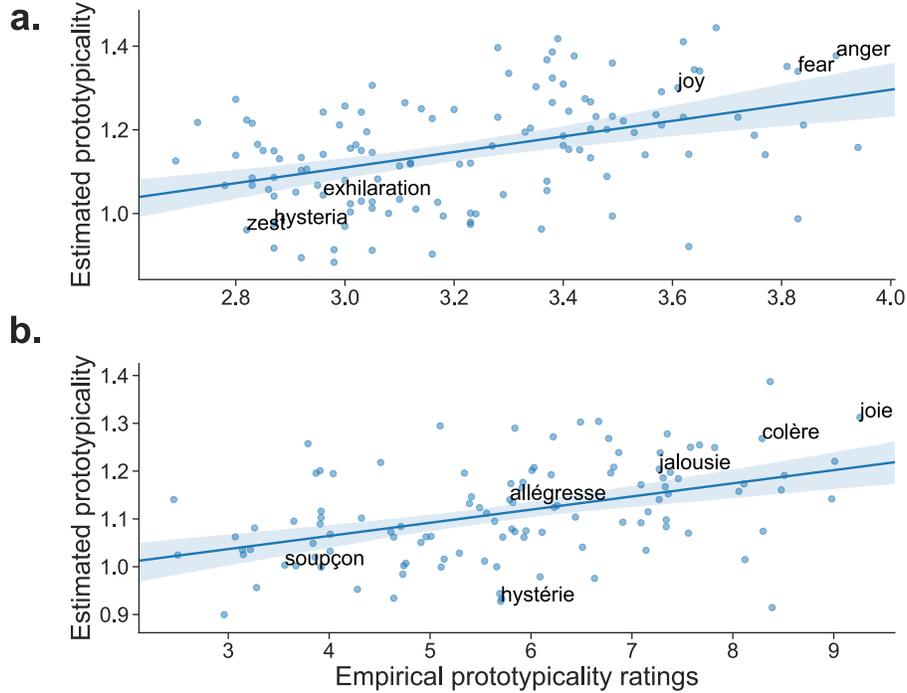}
\caption{Word embedding reconstruction of emotion prototypicality in a) English and b) French. Scatter plots compare estimated prototypicality computed from Equation~\ref{eq:2} against empirical ratings. Each dot corresponds to an emotion concept (a sample of concepts annotated), and each band shows a 95\% confidence interval for the line of best fit.} \label{fig:reconstruction}
\end{figure}

Figure~\ref{fig:reconstruction} shows the Pearson correlation between estimated prototypicality from English word embeddings and ratings from English speakers~\cite{shaver1987emotion}: $\rho = 0.428$, $p < 0.001$, $n = 123$. We obtained similar results with French word embeddings for a set of French emotion concepts~\cite{niedenthal2004prototype}: $\rho =0.438$, $p < 0.001$, $n = 112$. These initial results show our estimated degrees of prototypicality for emotion concepts capture human judgments to some extent. For this reason, we used the same method to estimate historical prototypicality ratings and evaluated them as predictors of semantic stability.

\begin{figure}[!ht]
\centering
\includesvg[width=\linewidth]{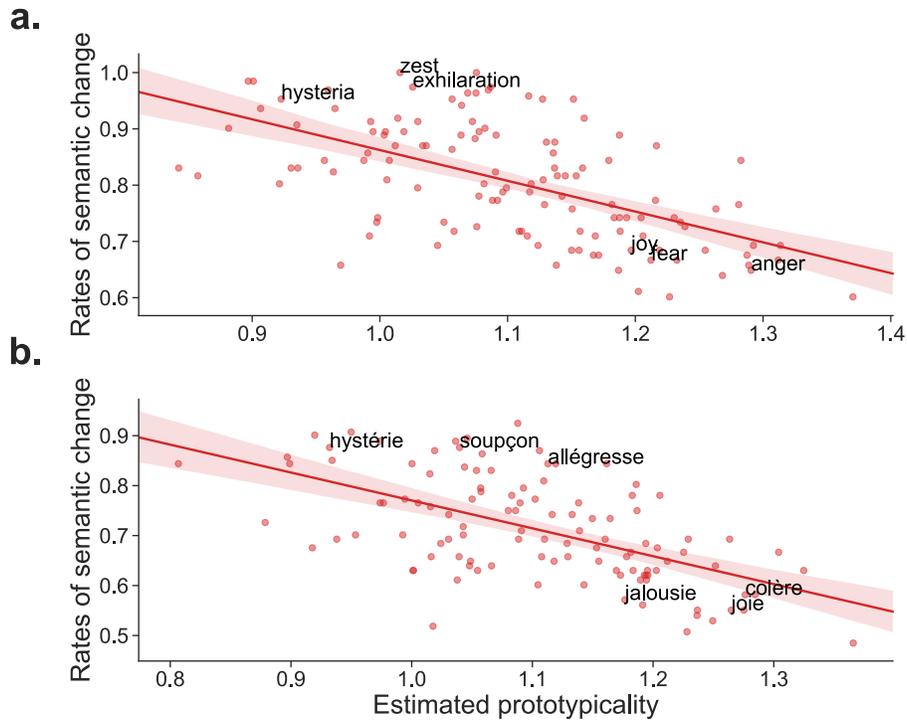}
\caption{Scatter plots showing the negative correlations between emotion prototypicality and rates of emotion semantic change between the 1890s and 1990s, in a) English and b) French. Each dot corresponds to an emotion word (with a sample set of words annotated), and each band shows a 95\% confidence interval for regressions between emotion prototypicality and rates of semantic change.} \label{fig:simple}
\end{figure}

Figure~\ref{fig:simple} shows a significant negative correlation between emotion prototypicality and degree of semantic change: $\rho = -0.580$, $p < 0.001$, $n = 123$. On average, emotion concepts rated prototypical such as {\it anger}, {\it joy}, {\it fear} underwent less change in meaning compared to words denoting less prototypical concepts such as {\it zest, exhilaration} and {\it hysteria} (see annotated word samples in Figure~\ref{fig:simple}). Similar results hold for French: $\rho=-0.576$, $p < 0.001$, $n = 112$. {\it Supplementary Information} provides additional examples of English and French emotion concepts from the most changing to the most semantically stable, along with their semantic neighbours retrieved from our methods.

\begin{figure}[!ht]
\centering
\includesvg[width=\linewidth]{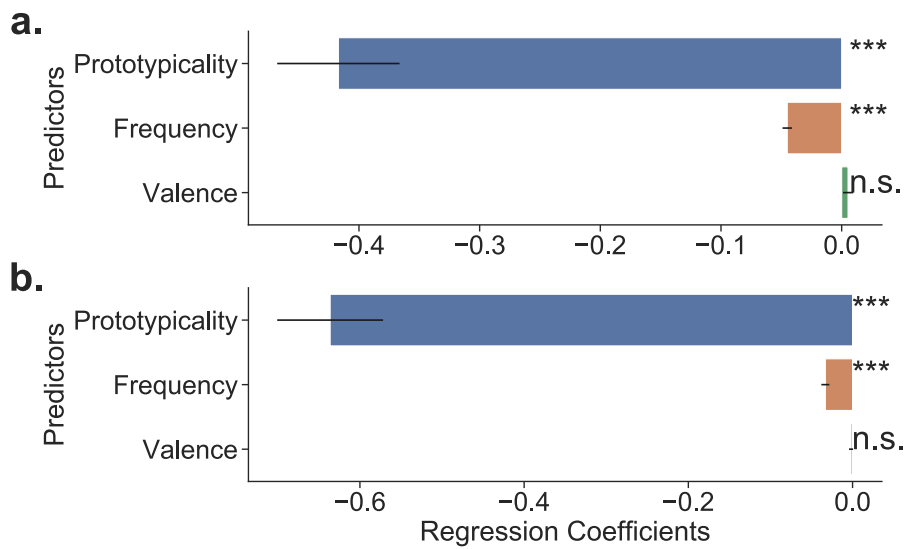}
\caption{Predictor coefficients from multiple regressions on rates of emotion semantic change. Error bars show standard error, and ``n.s.'', ``*'', ``**'', ``***'' denote no significance at $p<0.05$, and $p < 0.05$, $p<0.01$, $p<0.001$ respectively. a) shows results for English, and b) shows results for French.} \label{fig:multiple}
\end{figure}

Figure~\ref{fig:multiple} shows our results for multiple regression. The adjusted $R^2$ of the model for English is $0.680$, with $p < 0.001$, $n = 123$; mean regression coefficients for prototypicality ($\beta = -0.417$, $p < 0.001$) and frequency ($\beta = -0.0451$, $p <0.001$) are significant, but for valence ($\beta = 0.0053$, $p  = 0.208$) it is insignificant. For French, the adjusted $R^2$ of the model is $0.538$, with $p < 0.001$, $n = 112$; mean regression coefficients for prototypicality ($\beta = -0.6363$, $p < 0.001$) and frequency ($\beta = -0.0331$, $p < 0.001$) are significant, but for valence ($\beta = 0.0019$, $p = 0.454$) it is insignificant. These results show that frequency predicts semantic stability, which confirms the previous findings~\cite{hamilton2016diachronic,dubossarsky2017outta}. Beyond frequency, we find that  prototypicality plays an important role in predicting  semantic stability of emotion words, manifested in its significant and negative effect. This provides evidence for our hypothesis that prototypical emotion words tend to be semantically stable over time. 

{\it Supplementary Information} includes three more analyses that further corroborate our findings. The first analysis repeats the multiple regression but restricts the neighbourhoods to emotion concepts only when computing $rate(w,1890,1990)$; the results rule out the possibility that our findings are an artifact of the non-emotion senses of polysemous emotion concepts (e.g., {\it zest}). The second analysis extends the multiple regression for English by including additional predictors based on hypernymy-hyponymy, age of acquisition, and degrees of polysemy, which could potentially subsume the effects of prototypicality; our results show that this is not the case. The third analysis repeats the multiple regression for English emotion concepts, except the rates of change are computed as $rate(w,1980,1990)$ and empirical prototypicality from~\cite{shaver1987emotion} were used; these results provide evidence that the effect of prototypicality is not caused by potential artifacts in our estimation of prototypicality based on Equation~\ref{eq:2}. 

\begin{figure}[!ht]
\centering
\includesvg[width=\linewidth]{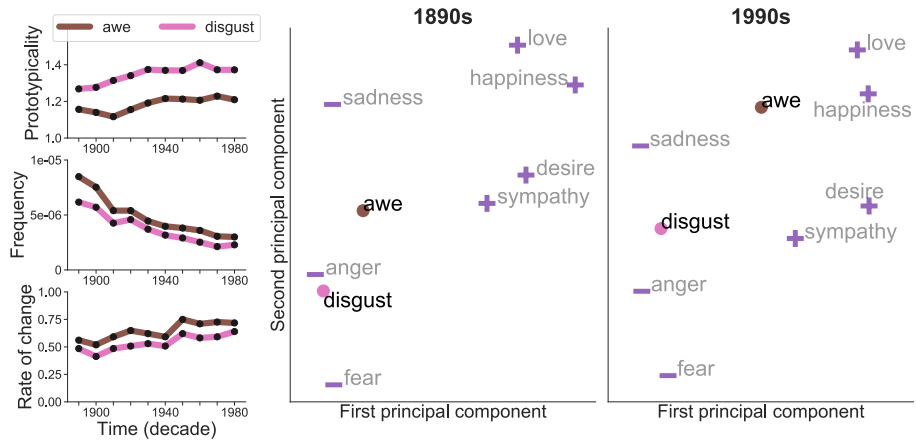}
\caption{An illustrative comparison of prototypicality, frequency, and  semantic stability in emotion words {\it awe} and {\it disgust}. Left panels show the embedding-based prototypicality, frequency and degree of semantic change of {\it awe} and {\it disgust} over time. Right panels visualize the rates of change in these words by placing them in the two principal components~\cite{wold1987principal} of meaning space, along side prototypical emotion concepts which are annotated based on their valence (``+'' for positive, ``-'' for negative).} \label{fig:illustration2}
\end{figure}

Figure~\ref{fig:illustration2} illustrates our main finding with two example words: {\it disgust} and {\it awe}. These words had similar usage frequencies over time, but {\it disgust} is rated as a more prototypical emotion word than {\it awe}~\cite{shaver1987emotion}. Over time, {\it awe} has shifted meaning more substantially than {\it disgust}. In particular, both words were in the neighbourhood of negative emotion words (e.g., {\it sadness}, {\it anger}, and {\it fear}) in the 1890s. However, while {\it disgust} still remained close to these words in the 1990s, {\it awe} moved closer to positive emotion words (e.g., {\it love} and {\it happiness}).

\section{Analyses of bird names}
In this set of analyses, we demonstrate that the sources of prototypicality do not always provide semantic stability as we have shown for emotion concepts. Here we repeat our previous analyses on a case study of birds, a frequently investigated category in prototype theory~\cite{rosch1975cognitive,rosch1978cognition}. As we will see, our embedding-based estimation of prototypicality does not work well with bird names, and we will focus our analysis on using empirical ratings from the 1970s.

\subsection{Materials}
We obtained a list of English bird names with prototypicality ratings from~\cite{rosch1975cognitive}. The author produced the list by consulting previous work so that the selected names were relatively frequent. They produced bird prototypicality ratings by asking 209 American university students to rate each of these names on a 7-point scale, where 1 means the name refers to a very good example of a bird, and 7 means the name refers to a very poor example. Note that the scale operates in the opposite direction of our prototypicality ratings for emotion concepts. For clarity, we multiplied these ratings by -1 so the direction is the same as our emotion data. Focusing on the 1970s and 1990s, we used historical data from HistWords~\cite{hamilton2016diachronic}, which was intersected with the word list and provided us with 41 bird names.

\subsection{Methods}
Similar to the previous section, we attempted at estimating bird prototypicality using Equation~\ref{eq:2}. We then computed the rates of change for every bird name $w$, $rate(w, 1970, 1990)$ using Equation~\ref{eq:1}. We computed the Pearson correlation between rates of change and prototypicality ratings obtained from the 1970s, and we performed a multiple regression using the following formula:
\begin{equation}
    rate(w, 1970, 1990) \sim proto(w) + freq(w) 
    \label{eq:4}
\end{equation}
where we denote the empirical prototypicality rating of every bird name $w$ as $proto(w)$.

\subsection{Results}
\begin{figure}[ht]
\centering
\includesvg[width=\linewidth]{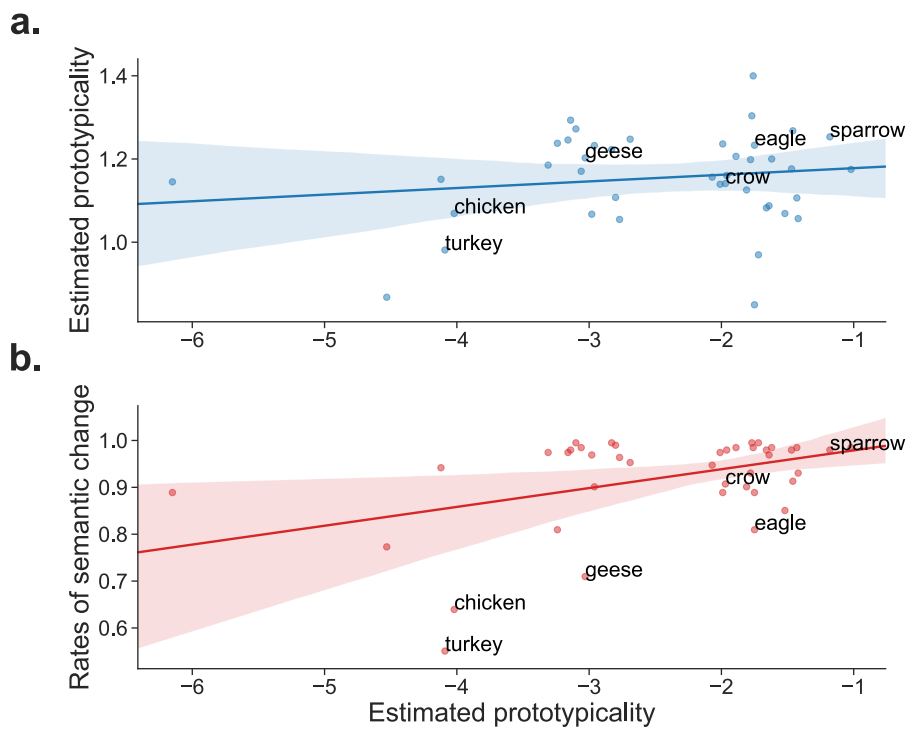}
\caption{Analyses of bird names: a) word embedding reconstruction of bird prototypicality and b) correlations between bird prototypicality and rates of semantic change between the 1970s and 1990s. Each dot corresponds to a bird name, and each band shows a 95\% confidence interval for the line of best fit.} \label{fig:bird_analyses}
\end{figure}

Figure~\ref{fig:bird_analyses}a shows the Pearson correlation between estimated prototypicality and empirical ratings from~\cite{rosch1975cognitive}: $\rho = 0.153$, $p = 0.340$, $n = 41$. While the same method reconstructs prototypicality for emotion concepts to some extent, our text-based method does not explain a significant amount of variance in the prototypicality of birds which depends more on sensory features~\cite{garrard2001prototypicality}. It has been shown that prototypical birds in our dataset tend to be passerines, small perching birds that sing (e.g., \textit{robin}), and less prototypical ones tend to be non-passerines (e.g., \textit{penguin})~\cite{boster1988natural}, which our text-based methodology did not capture. For this reason, we chose to focus on empirical prototypicality ratings for birds in our analyses.

\begin{figure}[ht]
\centering
\includesvg[width=\linewidth]{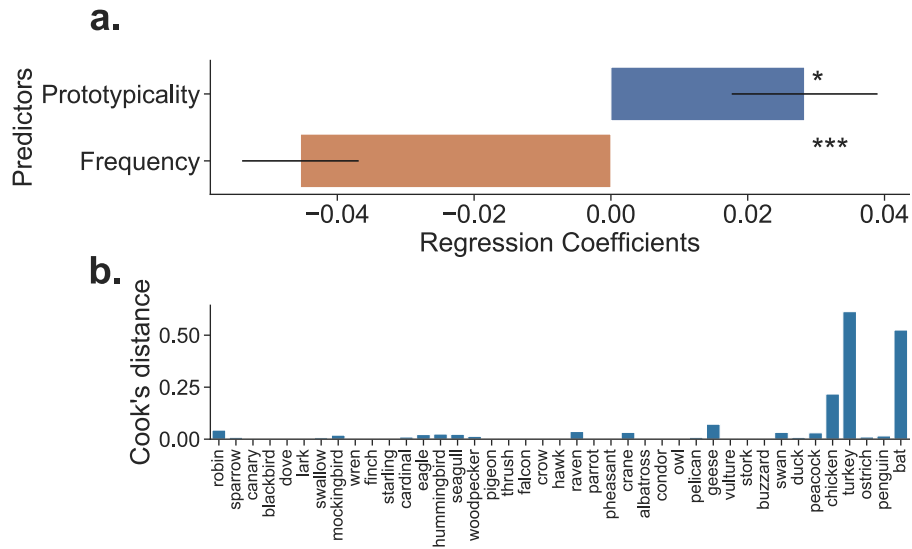}
\caption{Multiple regression analysis of bird names: a) predictor coefficients from multiple regressions on rates of semantic change, following the same layout as Figure~\ref{fig:multiple}; b) Cook's distance for every bird name, showing the influence of individual data points on the regression result.} \label{fig:multiple_birds}
\end{figure}

Figure~\ref{fig:bird_analyses}b shows a significant positive correlation between bird prototypicality and degree of semantic change: $\rho = 0.428$, $p = 0.005$, $n = 41$. This finding suggests that the relation between semantic change and prototypicality in bird names is opposite to our previous findings for emotion words. Figure~\ref{fig:multiple_birds}a shows the results for multiple regression. The adjusted $R^2$ is $0.508$, with $p < 0.001$, $n = 41$; mean regression coefficients for empirical prototypicality ($\beta = 0.0283$, $p = 0.011$) and frequency ($\beta = -0.0454$, $p < 0.001$) are significant. We observe frequency still predicts semantic stability, suggesting it is indeed a general predictor of semantic change. Interestingly, prototypicality of birds not only failed to predict stability as in the case of emotion concepts, but also pointed to the opposite trend: in the category of birds, names of prototypical birds tend to undergo more change than other names.

To better understand the implications of this variability to our finding about bird names, we performed a more in-depth analysis of the data. Unlike the case of emotion, we observe bird names exhibit high variability in Figure~\ref{fig:bird_analyses}b, which is reflected in the wide confidence region.\footnote{Note that the number of available bird terms for our analysis is substantially lower than that of emotion terms.} This suggests the opposite trend in bird names is influenced by only a handful of less prototypical birds. We estimated the influence of each bird name using Cook's distance, which takes into account the data point's residual and leverage. Figure~\ref{fig:multiple_birds}b shows the influence of each bird in the regression analysis: the most influential points are {\it turkey}, {\it bat} and {\it chicken}. We can observe {\it bat} is likely to be influential as it has a much higher rating (not prototypical) than other bird names; this might be because subjects in the original study, being university students, were familiar with the scientific classification of bats. More importantly, despite not being prototypical birds, {\it turkey} and {\it chicken} could have important cultural roles (festive or culinary) in North America so that they provided anchors for their meaning, thereby contributing to the significant correlation between bird prototypicality and semantic change.

\begin{figure}[ht]
\centering
\includesvg[width=\linewidth]{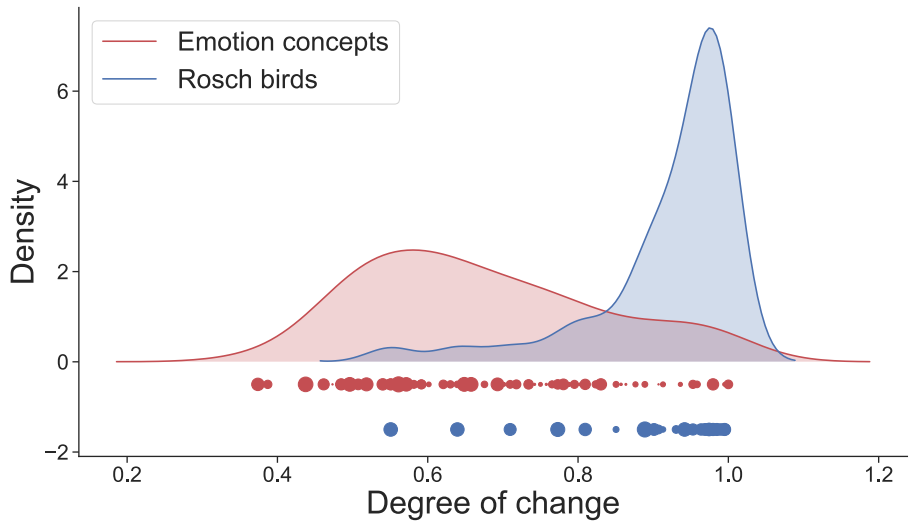}
\caption{Distributions of semantic change in emotion and bird categories. Each dot corresponds to a word, and the size of the dot is proportional to its degree of prototypicality. The density plots were obtained using kernel density estimation; although degrees of change given by Equation~\ref{eq:1} are technically bounded between 0 and 1, we did not bound the support of this figure for illustrative reasons.} \label{fig:compare}
\end{figure}

Figure~\ref{fig:compare} compares the degrees of semantic change that took place in emotion concepts and bird names between the 1970s and 1990s. Many prototypical emotion concepts tend to lie at the lower tail of the density distribution and show high stability, mirroring the results we have seen previously, but the same pattern does not hold for birds. We observe that overall bird names tend to undergo greater change than emotion concepts do. It is possible that prototypical birds possess the most representative features of the bird category, which could provide points of attachment for meaning change via processes such as chaining, in which a word for one object is extended to be used for another, or metaphor~\cite{luo2018stability,ramiro2018algorithms}.  This general pattern of more rapid change among bird names together with the additional semantic stability of a handful of influential bird exemplars may be responsible for the positive correlation between degrees of bird prototypicality and rates of semantic change. 

\section{Conclusion}
Language offers a lens into the history of emotion semantics. Our computational linguistic analyses of semantic change suggest that a new view of emotion concepts in language may be warranted. Rather than perceiving of emotion concepts as static, their meaning is evolving over time. The exact cultural or societal factors responsible for semantic change in emotion words are difficult to pinpoint, and they may be different for each emotion term. For example, semantic change in \textit{awe} may reflect a movement away from its use in religious contexts, in which it reflects more of a fearful respect, towards greater use in beautiful artistic and natural contexts that followed the emergence of romanticism and transcendentalism in the early to middle nineteenth century. We assessed semantic change over a relatively short timescale, suggesting that in the centuries to come it is possible that words like \textit{awe} may continue to evolve and mean something very different than they do today. 

Further, we found in two languages that more prototypical emotion words~\cite{shaver1987emotion,rosch1978cognition} showed greater semantic stability than other emotion words over time. The relation between prototypicality and semantic change depends on its exact sources, as we observed opposite trends for emotions and birds. The importance of prototypicality as a predictor in semantic change for other semantic categories remains an open question and future work should investigate what features affect the importance of prototypicality. Our study extends research on emotions to its historical development and offers a computational cognitive characterization of evolving emotion semantics from natural language use.

\section*{Acknowledgements}
We thank Brett Ford and two anonymous reviewers for providing constructive comments on our manuscript. This work was supported by a University of Toronto Graduate Entrance Scholarship to AX, a Canada Foundation for Innovation grant \#502426 to JES, a NSERC Discovery Grant RGPIN-2018-05872 and a SSHRC Insight Grant \#435190272 to YX.

\appendix
\section{Supplementary material}

Code and data used for our analyses are available on GitHub at \url{https://github.com/johnaot/Emotion_Semantic_Change}.

\section{Word age and prototypicality}
Here we further demonstrate the differences between prototypicality in emotions and prototypicality in birds. Previous work has suggested that prototypical emotion concepts are well-defined and may have particularly strong social or cultural scripts~\cite{barrett2006emotions,fehr1984concept,russell1999core}. Thus it is conceivable that words for prototypical emotion concepts exist in a lexicon prior to those less prototypical emotion words. In contrast, prototypicality of birds is based on biological taxonomy~\cite{boster1988natural} and grounded in sensory and visual perception~\cite{garrard2001prototypicality}, and we do not expect prototypicality to be reflected in the age of a word; it is likely that a relatively newly documented passerine (e.g., bluebird) entered the lexicon after well-established non-passerines (e.g., chicken). To test these ideas, we analyzed the correlation between word age and prototypicality in the categories of emotion and birds.

Following existing work~\cite{xu2019wordforms}, we obtained the age of a word from the Historical Thesaurus of English (HTE)~\cite{HTE4.2}. For each word entry, the HTE provides a list of senses of the word, and for each sense, the HTE provides the word class associated with that sense of the word and the date of first appearance of the sense in historical records. We operationalized the date of (the first) emergence of a word to be the earliest date among the dates of first appearance across all of its senses. Since our analyses focused on nouns, we considered only noun senses. 
We did not analyze the age of French words due to the unavailability of comparable French dictionaries.

\begin{figure}[!ht]
\centering
\includesvg[width=\linewidth]{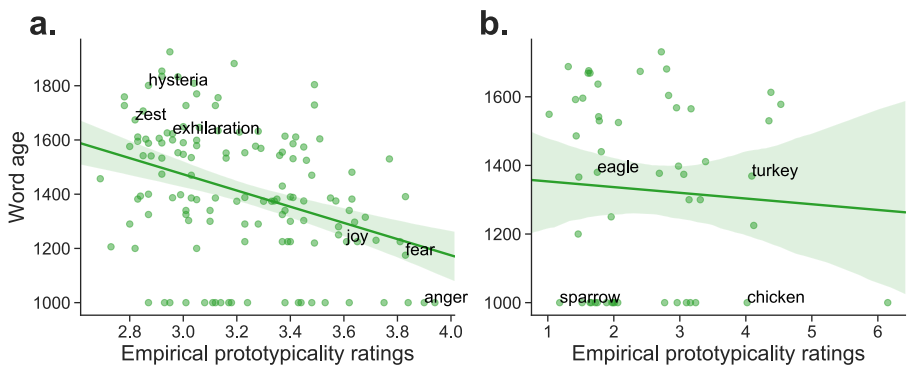}
\caption{Scatter plots showing the relations between prototypicality and word age for a) emotion words and b) bird names. Each dot corresponds to a word (with a sample set of words annotated), and each band shows a 95\% confidence interval for regressions between prototypicality and age.} \label{fig:age}
\end{figure}

We analyzed the same lists of English emotion words and bird names described in {\it Section 3.1} and {\it Section 4.1}, which we intersected with the HTE data. For English emotion words, the Pearson correlation between emerging date and prototypicality is $-0.366$, $p < 0.001$, $n = 135$. This indicates that prototypical emotion words emerged earlier in the history of English. The same pattern does not hold for the bird category. For Rosch bird names, the Pearson correlation between emergence date and prototypicality is $-0.0657$, $p = 0.643$, $n = 52$. See Figure~\ref{fig:age} for illustration.

\section{Evaluation of the nearest-neighbour measure}
We show that the nearest-neighbour measure of semantic change described in {\it Section 2} is 1) robust to variation in neighbourhood size (denoted by $k$), and 2) interpretive for a word's semantic change based on nearest neighbours retrieved at different time points.

\subsection{Robustness in neighbourhood size $k$}
\begin{figure}[!ht]
\centering
\includesvg[width=\linewidth]{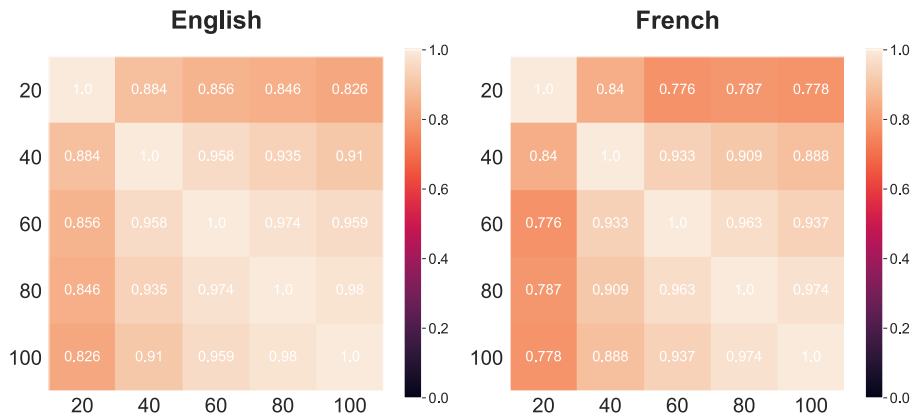}
\caption{Robustness of $kNN$ demonstrated using emotion words. The first column shows results for English and the second column shows results for French. Each cell shows the Pearson correlation between changes measured by $x$-nearest neighbours and by $y$-nearest neighbours. All p-values are significant and less than 0.001.}
\label{fig:robust_k_emotion}
\end{figure}

We evaluate whether the nearest-neighbour measure is robust to variation in $k$. Following {\it Section 3.1}, we again quantify semantic change by setting $t_1 = 1890$ and $t_2 = 1990$ and focus our analysis on the lists of English and French emotion words we analyzed in the main text. We show that for $k = 20, 40, 60, 80,$ and $100$, resulting degrees of semantic change $rate(w, t_1, t_2)$ are highly correlated. The correlation results are summarized in Figure~\ref{fig:robust_k_emotion}. We observed that in both English and French, the degrees of change are strongly correlated between any two of the predetermined settings of $k$, with a small decrease in correlations for the lowest value of $k=20$.

\subsection{Examples of emotion semantic change}
Qualitative changes in the nearest neighbours of a word offer interpretability for semantic change. We provide examples of emotion words that underwent the most and the least changes and their nearest semantic neighbours in Tables~\ref{table:example_english_most},~\ref{table:example_english_least},~\ref{table:example_french_most}, and~\ref{table:example_french_least}. For example, in Table~\ref{table:example_english_most}, we observed {\it zest}, which used to primarily convey joy but later became primarily associated with food, is among the most changing emotion words; on the other hand, in Table~\ref{table:example_english_least}, we observed words like {\it surprise} has barely changed in meaning. Similarly, in Table~\ref{table:example_french_most}, we observed {\it stupéfaction} in French become less associated with despair and anxiety over time; in Table~\ref{table:example_french_least}, the least changing words on the other hand tend to preserve similar emotion words as neighbours.

\begin{table*}[!ht]
\centering
\scriptsize
 \begin{tabular}{||l | l | l||} 
 \hline
 Most Changing & Nearest Neighbours in 1890s & Nearest Neighbours in 1990s \\
 \hline\hline
 zest & relish, enjoyment, sprightliness & juice, teaspoons, vinegar \\ 
 \hline
 infatuation & priestcraft, devastations, misanthrope & inhomogeneity, palates, pleurisy \\
 \hline
 sentimentality & cant, sentimentalism, rusticity & polyphony, sterne, mandel \\ 
 \hline
 optimism & pessimism, aptness, sentimentalism & pessimism, insecurity, enthusiasm \\ 
 \hline
 exhilaration & mountebank, festivity, tulip & joy, sadness, excitement \\ 
 \hline
 aggravation & misery, symptoms, consequences & proxies, sleeplessness, stressor \\
 \hline
 exasperation & vehemence, peevishness, pitch & astonishment, amazement, disgust \\
 \hline
 glee & merriment, wassail, delight & claps, shouts, megaphone \\
 \hline
 cheerfulness & hopefulness, sprightliness, vivacity & serenity, blasphemies, cannonade \\
 \hline
 gaiety & sprightliness, vivacity, gayety & pontellier, faints, plaudits \\
 \hline
 fondness & liking, peevishness, passion & affection, pianists, groanings \\
 \hline
 hysteria & neurasthenia, simulation, melancholia & neurosis, hypochondriasis, psychosis \\
 \hline
 dejection & despondency, sullenness, irresolution & pantomime, theseus, disquiet \\
 \hline
 elation & sullenness, despondency, peevishness & despair, revulsion, dread \\
 \hline
 ferocity & fierceness, cruelty, prowess & vigor, fury, proverb \\
 \hline
 revulsion & feeling, disquietude, outburst & disgust, hisses, yearnings \\
 \hline
 isolation & loneliness, disorganization, seclusion & monger, characterization, coli \\
 \hline
 alienation & eviction, property, repugnancy & helplessness, blauner, resentment \\
 \hline
 hopelessness & uselessness, futility, helplessness & helplessness, despair, frustration \\
 \hline
 rapture & ecstasy, delight, joy & joy, indignation, outcasts \\
 \hline
\end{tabular}
\caption{Top 20 most changing English emotion words along with their 3 nearest neighbours in the flanking decades.}
\label{table:example_english_most}
\end{table*}

\begin{table*}[!ht]
\centering
\scriptsize
 \begin{tabular}{||l | l | l||} 
 \hline
 Least Changing & Nearest Neighbours in 1890s & Nearest Neighbours in 1990s \\
 \hline\hline
 grief & sorrow, anguish, joy & sorrow, sadness, anguish \\ 
 \hline
 pity & compassion, love, sympathy & compassion, shame, sadness \\ 
 \hline
 misery & wretchedness, miseries, degradation & sorrow, bitterness, anguish \\ 
 \hline
 disgust & horror, aversion, indignation & sadness, annoyance, amazement \\ 
 \hline
 anger & indignation, resentment, rage & resentment, rage, frustration \\
 \hline
 surprise & astonishment, amazement, dismay & astonishment, amazement, dismay \\ 
 \hline
 sorrow & grief, anguish, sadness & grief, sadness, misery \\
 \hline
 affection & affections, tenderness, esteem & admiration, sympathy, love \\
 \hline
 happiness & felicity, prosperity, welfare & prosperity, joy, enjoyment \\
 \hline
 despair & desperation, dismay, rage & anguish, frustration, sadness \\
 \hline
 fear & dread, anger, shame & dread, resentment, anger \\
 \hline
 horror & terror, astonishment, amazement & terror, astonishment, amazement \\
 \hline
 regret & disappointment, grief, sorrow & disappointment, sadness, bitterness \\
 \hline
 envy & jealousy, uncharitableness, hatred & jealousy, hatred, resentment \\
 \hline
 disappointment & mortification, grief, regret & frustration, sadness, regret \\
 \hline
 rage & fury, anger, indignation & anger, fury, indignation \\
 \hline
 shame & disgrace, infamy, blush & guilt, pity, humiliation \\
 \hline
 astonishment & amazement, surprise, dismay & amazement, dismay, surprise \\
 \hline
 joy & gladness, delight, grief & delight, sorrow, excitement \\
 \hline
 sympathy & sympathies, compassion, affection & affection, admiration, compassion \\
 \hline
\end{tabular}
\caption{Top 20 least changing English emotion words along with their 3 nearest neighbours in the flanking decades.}
\label{table:example_english_least}
\end{table*}

\begin{table*}[!ht]
\centering
\scriptsize
 \begin{tabular}{||l | l | l||} 
 \hline
 Most Changing & Nearest Neighbours in 1890s & Nearest Neighbours in 1990s \\
 \hline\hline
 stupéfaction & indiscrétion, désespoir, anxiété & désapprobation, émotion, allégresse \\ 
 \hline
 suspicion & inculpation, défiance, prévention & méfiance, défiance, incertitude \\
 \hline
 culpabilité & réussite, identité, présomption & infériorité, persécution, châtiment \\ 
 \hline
 déplaisir & étonnement, inquiétude, appréhensions & plaisir, mâle, océans \\ 
 \hline
 torpeur & engourdissement, apathie, léthargie & apathie, consternation, stupeur \\
 \hline
 extase & contemplation, stupeur, somnambulisme & contemplation, joie, angoisse \\ 
 \hline
 soupçon & soupçons, équivoque, délit & préjugé, partialité, préjugés \\
 \hline
 hystérie & épilepsie, diabète, étiologie & névrose, épilepsie, névroses \\
 \hline
 séduction & adultère, entraînements, cruauté & immédiateté, impiété, éloquence \\
 \hline
 désolation & épouvante, dévastation, misère & pauvreté, saleté, nausées \\
 \hline
 froideur & bonhomie, trousseau, bassesse & arrogance, ingratitude, insolence \\
 \hline
 excitation & irritation, nerfs, nerf & tension, angoisse, agitation \\
 \hline
 intimidation & violence, ruse, corruption & menaces, coercition, chantage \\
 \hline
 timidité & fierté, délicatesse, naïveté & docilité, avidité, découragement \\
 \hline
 tension & volts, pression, potentiel & appareillage, excitation, tensions \\
 \hline
 intérêt & intérêts, utilité, équité & intérêts, utilité, rentabilité \\
 \hline
 espérance & espoir, espérances, désir & espoir, espérances, désir \\
 \hline
 dépit & mépris, défaillances, hésitations & grâce, précocité, conséquence \\
 \hline
 allégresse & fierté, épouvante, gaieté & joie, émotion, gaieté \\
 \hline
 effusion & sang, tendresse, larmes & sang, amertume, lucre \\
 \hline
\end{tabular}
\caption{Top 20 most changing French emotion words along with their 3 nearest neighbours in the flanking decades.}
\label{table:example_french_most}
\end{table*}

\begin{table*}[!ht]
\centering
\scriptsize
 \begin{tabular}{||l | l | l||} 
 \hline
 Least Changing & Nearest Neighbours in 1890s & Nearest Neighbours in 1990s \\
 \hline\hline
 tristesse & angoisse, amertume, effroi & amertume, douleur, angoisse \\ 
 \hline
 tendresse & bienveillance, sympathie, sollicitude & douceur, compassion, amour \\ 
 \hline
 patience & courage, persévérance, prudence & courage, persévérance, audace \\ 
 \hline
 orgueil & vanité, amour, ambition & arrogance, insolence, vanité \\ 
 \hline
 horreur & honte, effroi, angoisse & opprobre, effroi, tristesse \\ 
 \hline
 effroi & tristesse, consternation, terreur & tristesse, consternation, horreur \\
 \hline
 indignation & admiration, effroi, cri & enthousiasme, cri, admiration \\
 \hline
 joie & tristesse, douleur, bonheur & tristesse, enthousiasme, douleur \\
 \hline
 honte & horreur, chagrin, humiliation & opprobre, humiliation, peur \\
 \hline
 jalousie & haine, ambition, convoitise & arrogance, haine, rancune \\
 \hline
 colère & désespoir, mécontentement, anxiété & désespoir, fureur, émotion \\
 \hline
 douleur & chagrin, souffrance, douleurs & souffrance, tristesse, douleurs \\
 \hline
 stupeur & tristesse, angoisse, effroi & consternation, surprise, effroi \\
 \hline
 vengeance & haine, ressentiment, fureur & haine, jalousie, orgueil \\
 \hline
 bonheur & malheur, gloire, joie & malheur, joie, plaisir \\
 \hline
 chagrin & douleur, tristesse, honte & tristesse, douleur, amertume \\
 \hline
 terreur & effroi, horreur, haine & horreur, effroi, anarchie \\
 \hline
 enthousiasme & ardeur, joie, admiration & joie, admiration, indignation \\
 \hline
 impatience & anxiété, indignation, angoisse & joie, humeur, espoir \\
 \hline
 souffrance & douleur, tristesse, angoisse & douleur, angoisse, souffrances \\
 \hline
\end{tabular}
\caption{Top 20 least changing French emotion words along with their 3 nearest neighbours in the flanking decades.}
\label{table:example_french_least}
\end{table*}

\section{Additional analyses of emotion semantic change}
We describe three additional analyses that corroborate our findings on emotion semantic change in the main text. The first analysis rules out the possibility that our findings are an artifact of the non-emotion senses of polysemous emotion concepts (e.g., {\it zest}). The second analysis shows that additional predictors based on hypernymy-hyponymy and degrees of polysemy do not subsume the effects of prototypicality on emotion semantic change. The third analysis provides evidence that our results on emotion concepts were not caused by artifacts in our estimation of prototypicality.


\subsection{Category-bounded analysis}
We investigate the robustness of our semantic change analyses by considering a variant of the nearest-neighbour measure discussed in {\it Section 2}. Originally in {\it Section 2}, the degree of semantic change of a word $w$ is defined as the Jaccard distance between its nearest neighbours at time $t_1$, $kNN(t_1)$, and its nearest neighbours at a later time, $kNN(t_2)$, where $kNN$ is restricted to nouns in the entire lexicon and determined by cosine similarity over word vectors. Since the meaning of every emotion word is represented by one word vector only, the set $kNN$ might also capture meaning change with respect to the word's polysemes and homonyms, i.e., meaning change outside the category of emotion. To assess how such meaning change might affect our results, we restricted the set of nearest neighbours so that only the list of emotion words are included, i.e., a category-bounded analysis of emotion semantic change. Since the set of emotion words is much smaller than the full lexicon, we set the size of the neighbourhood to be $k = 25$. 

We first provide evidence that this variant of the nearest-neighbour measure is also capable of capturing semantic change by showing 1) this measure is positively correlated with the original nearest-neighbour measure, and 2) this measure captures the negative relationship between frequency and semantic change~\cite{hamilton2016diachronic}. We obtain degrees of change under this variant measure by following the same procedure described in {\it Section 2}. In the case of English emotion words, the Pearson correlation between degrees of semantic change measured by this variant and degrees obtained by the original measure is $0.751$, $p < 0.001$, $ n = 123$; the Pearson correlation between degrees of change measured by the variant and frequency is $-0.489$, $p < 0.001$, $ n = 123$. In the case of French emotion words, the Pearson correlation between degrees of change measured by this variant and degrees obtained by the original measure is $0.604$, $p < 0.001$, $ n = 112$; the Pearson correlation between degrees of change measured by the variant and frequency is $-0.203$, $p = 0.0318$, $ n = 112$. These results suggest this variant is capable of replicating patterns of change identified previously. 

\begin{figure}[!ht]
\centering
\includesvg[width=\linewidth]{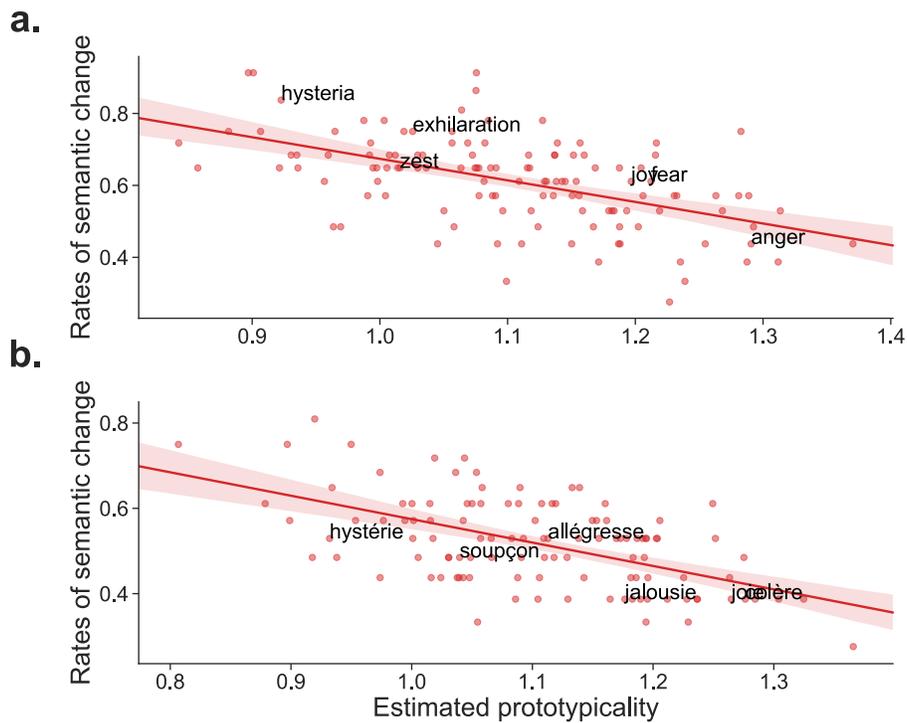}
\caption{Scatter plots showing the negative correlations between emotion prototypicality and rates of semantic change between the 1890s and 1990s, in a) English and b) French. Each dot corresponds to an emotion term (with a sample set of words annotated), and each band shows a 95\% confidence interval for regressions between prototypicality and rates of semantic change.}
\label{fig:bounded_emotion_scatter}
\end{figure}

\begin{figure}[!ht]
\centering
\includesvg[width=\linewidth]{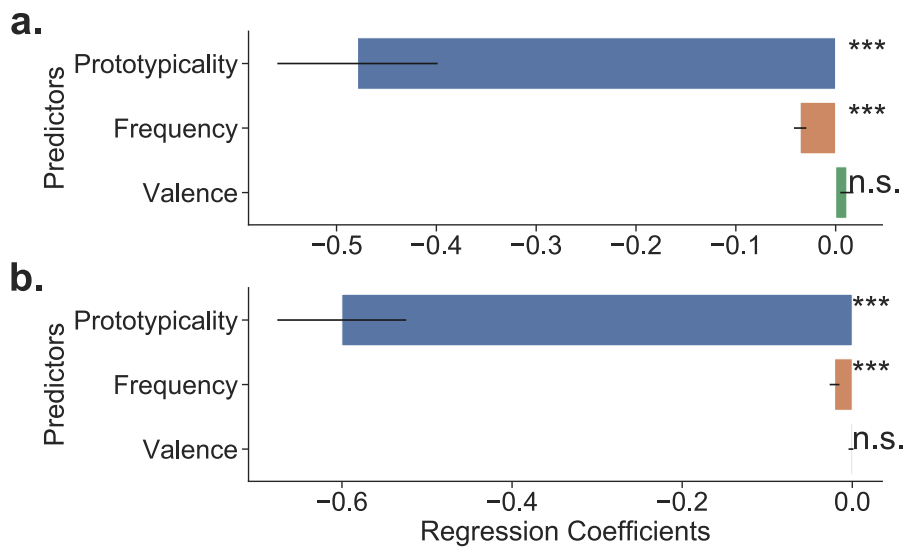}
\caption{Predictor coefficients from multiple regressions on rates of semantic change. Error bars show standard error, and ``n.s.'', ``*'', ``**'', ``***'' denote no significance at $p<0.05$, and $p < 0.05$, $p<0.01$, $p<0.001$ respectively. a) shows results for English, and b) shows results for French.}
\label{fig:bounded_emotion_multiple}
\end{figure}

After validating this variant measure, we also repeated the analyses on emotion semantic change described in the main text. Figure~\ref{fig:bounded_emotion_scatter} shows a significant negative correlation between prototypicality and degree of semantic change: for English, $\rho = -0.535$, $p < 0.001$, $n = 123$; for French, $\rho=-0.558$, $p = 0.002$, $n = 112$. Figure~\ref{fig:bounded_emotion_multiple} shows multiple regression results similar to the results presented in the main text. The adjusted $R^2$ for English is 0.432, with $p < 0.001$, $n = 123$. Mean regression coefficients for prototypicality ($\beta = -0.479$, $p < 0.001$) and frequency ($\beta = -0.0356$, $p <0.001$) remained negative and significant, whereas valence ($\beta=0.0112$, $p = 0.091$) is insignificant. Again, results hold similarly for French with the adjusted $R^2 = 0.381$, $p < 0.001$, $n = 112$ (prototypicality $\beta=-0.600$, $p < 0.001$; frequency $\beta=-0.0208$, $p < 0.001$; valence $\beta = -0.0015$, $p=0.599$). Compared to the main results, we observed that prototypicality remains a competitive predictor of semantic stability relative to frequency.

\subsection{Other factors of semantic change}
In the main text, we tested prototypicality as a predictor of semantic stability alongside frequency. Here we examine the role of prototypicality in predicting semantic stability by controlling for three additional predictors: 1) the degree of polysemy of a word, 2) superordinate-subordinate relations between emotion words, and 3) the age of acquisition (AoA) of a word. Firstly, similar to frequency, we control for degree of polysemy because it is a general predictor of semantic change which has been found to negatively correlate with stability in meaning~\cite{hamilton2016diachronic,luo2018stability}. Secondly, since one function of prototypical emotion words is that they can help define more complex emotion words~\cite{johnson1989language} and  this anchoring function may provide relative semantic stability,\footnote{For example, suppose {\it joy} is part of the definition of {\it exhilaration}; if the meaning of {\it joy} changed, the meaning of {\it exhilaration} will necessarily change as well, but not vice versa.} we examine superordinate-subordinate relations (i.e., the hierarchy in the semantic category of emotion) as a potential confounding variable influencing both prototypicality and semantic change. Thirdly, since prototypical emotion words are relatively well-defined and would be easy to learn, and that AoA is a known predictor of stability in lexical change~\cite{monaghan2014age,monaghan2019cognitive}, we control for AoA as a potential mediator between prototypicality and semantic change.

Following~\cite{luo2018stability}, we operationalized the degree of polysemy of a word as the number of senses the word had at the starting time $t_1 = 1890$ according to the HTE~\cite{HTE4.2}. To operationalize superordinate-subordinate relations, we used WordNet~\cite{fellbaum1998} provided by NLTK~\cite{nltk}. Specifically, we constructed a directed graph based on hypernym-hyponym relations, where the root is the sense for {\it feeling}, and the other nodes correspond to the most frequent sense of an emotion word (see Figure~\ref{fig:wn} for illustration). Then, we quantified a word's degree of subordination as its depth in the graph. For example, the word {\it thrill} has a depth of 4, while {\it joy} has a depth of 2. Furthermore, to match the historical period of our analyses, we used objective, test-based measurements of AoA originally published by~\cite{dale1981living}. A digitized version of this data was obtained from~\cite{brysbaert2017test}, where each entry contains a word form, its meaning, and the age at which it was acquired. We computed the AoA of a word by taking the average over all entries in which it appears.  Due to the lack of analogous French historical data, we only focused on English emotion words. We assumed these hypernym-hyponym relations and AoA are stable over the past century. After intersecting WordNet and AoA with our historical resources described in the main text, we had a total of 109 English emotion words.

\begin{figure}[!ht]
\centering
\includesvg[width=\linewidth]{other_factors/wn.svg}
\caption{WordNet hierarchy of hypernyms and hyponyms for a) positive emotion words and b) negative emotion words. Valence is determined by our data described in {\it Section 3.1}.} \label{fig:wn}
\end{figure}

We analyzed these factors using our materials and methods described in {\it Section 3.1} and {\it Section 3.2} of the main text. Specifically, we computed the rates of change for every emotion concept $x$, $rate(x, 1890, 1990)$. Then, we performed multiple regression using the following formula:
\begin{align}
    rate(x, 1890, 1990) \sim &p(x | c = emotion) + freq(x) + poly(x) + \\
    &val(x) + depth(x) + AoA(x)
    \label{eq:other_factors_regression}
\end{align}
where $poly(x)$ is the degree of polysemy of the word operationalized by number of senses, $depth(x)$ is the depth of the concept in the hypernym-hyponym graph we constructed, and $AoA(x)$ is the age at which the word was acquired. 

\begin{figure}[!ht]
\centering
\includesvg[width=\linewidth]{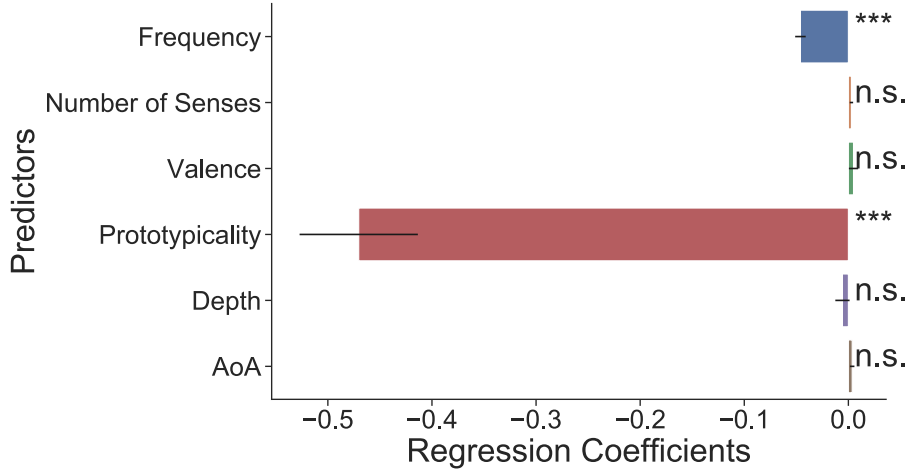}
\caption{Predictor coefficients from multiple regressions on rates of semantic change. Error bars follow the same layout as Figure~\ref{fig:bounded_emotion_multiple}. Prototypicality is estimated using Equation 2 in the main text.} \label{fig:other_factors}
\end{figure}

Figure~\ref{fig:other_factors} shows the multiple regression results suggesting the dominant roles of prototypicality and frequency. The adjusted $R^2$ of the model is $0.697$, with $p < 0.001$, $n = 109$; mean regression coefficients for prototypicality ($\beta = -0.4704$, $p < 0.001$) and frequency ($\beta = -0.0460$, $p <0.001$) are significant, but for valence ($\beta = 0.0049$, $p = 0.267$), number of senses ($\beta= 0.0029$, $p = 0.075$), depth ($\beta= -0.0057$, $p = 0.408$), and AoA ($\beta= 0.0036$, $p = 0.093$) it is insignificant. We observe that prototypicality still has a significant, negative effect as predicted by our hypothesis. We also observe that we can reproduce the finding by~\cite{hamilton2016diachronic} for frequency.

\begin{figure}[!ht]
\centering
\includesvg[width=\linewidth]{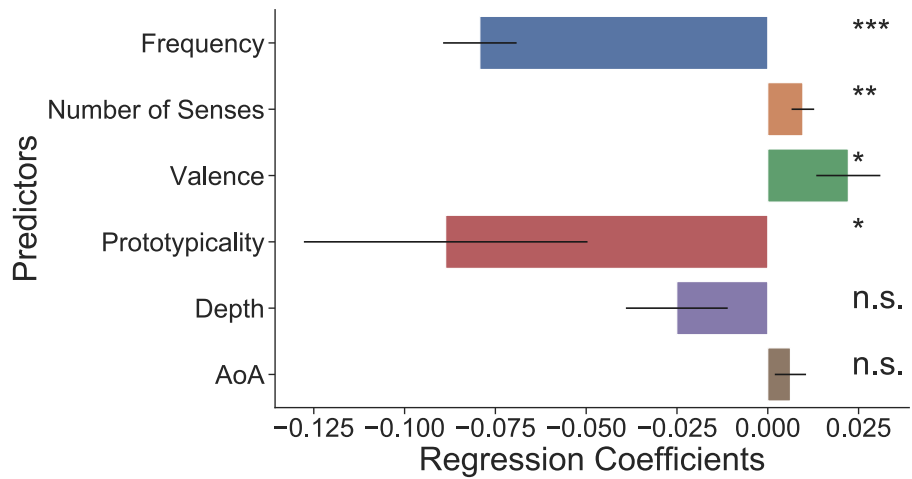}
\caption{Predictor coefficients from multiple regressions on rates of semantic change. Error bars follow the same layout as Figure~\ref{fig:bounded_emotion_multiple}. Prototypicality is based on human ratings.} \label{fig:empirical}
\end{figure}

\subsection{Human judgements of prototypicality}
We repeated the analysis in the previous section by replacing estimated prototypicality with empirical prototypicality ratings. We computed the rates of change for every emotion concept $x$, $rate(x, 1980, 1990)$. Then, we performed a multiple regression using the following formula:
\begin{equation}
    rate(x, 1980, 1990) \sim proto(x) + freq(x) + poly(x) + val(x) + depth(x)
    \label{eq:human_judgement_regression}
\end{equation}
where $proto(x)$ is the prototypicality rating of $x$ obtained from~\cite{shaver1987emotion}.

Figure~\ref{fig:empirical} shows the multiple regression results. The adjusted $R^2$ of the model is $0.524$, with $p < 0.001$, $n = 109$; mean regression coefficients for prototypicality ($\beta = -0.0887$, $p = 0.025$), frequency ($\beta = -0.0793$, $p < 0.001$), number of senses ($\beta = 0.0097$, $p = 0.002$), and valence ($\beta = 0.0222$, $p = 0.014$) are significant, but for depth ($\beta= -0.0250$, $p = 0.077$) and AoA ($\beta = 0.0062$, $p = 0.150$) it is insignificant. We observe that empirical prototypicality also has a significant, negative effect, albeit the effect size is smaller than previously.

\bibliography{elsarticle-template}

\begin{thebibliography}{10}
\expandafter\ifx\csname url\endcsname\relax
  \def\url#1{\texttt{#1}}\fi
\expandafter\ifx\csname urlprefix\endcsname\relax\def\urlprefix{URL }\fi
\expandafter\ifx\csname href\endcsname\relax
  \def\href#1#2{#2} \def\path#1{#1}\fi

\bibitem{kitayama1994emotion}
S.~E. Kitayama, H.~R.~E. Markus, Emotion and culture: {E}mpirical studies of
  mutual influence, American Psychological Association, 1994.

\bibitem{wierzbicka1999emotions}
A.~Wierzbicka, Emotions across Languages and Cultures: {D}iversity and
  Universals, Cambridge University Press, 1999.

\bibitem{jackson2019emotion}
J.~C. Jackson, J.~Watts, T.~R. Henry, J.-M. List, R.~Forkel, P.~J. Mucha, S.~J.
  Greenhill, R.~D. Gray, K.~A. Lindquist, Emotion semantics show both cultural
  variation and universal structure, Science 366~(6472) (2019) 1517--1522.

\bibitem{keltner2003approach}
D.~Keltner, J.~Haidt, Approaching awe, a moral, spiritual, and aesthetic
  emotion, Emotion and Cognition 17~(2) (2003) 297--314.

\bibitem{fehr1984concept}
B.~Fehr, J.~A. Russell, Concept of emotion viewed from a prototype perspective,
  Journal of Experimental Psychology: General 113~(3) (1984) 464.

\bibitem{shaver1987emotion}
P.~Shaver, J.~Schwartz, D.~Kirson, C.~O'connor, Emotion knowledge: {F}urther
  exploration of a prototype approach., Journal of Personality and Social
  Psychology 52~(6) (1987) 1061.

\bibitem{cosmides2000evolutionary}
L.~Cosmides, J.~Tooby, Evolutionary psychology and the emotions, in: M.~Lewis,
  H.-J.~J. M (Eds.), Handbook of Emotions 2nd Edition, Guilford, NY, 2000, pp.
  91--115.

\bibitem{barrett2006emotions}
L.~F. Barrett, Are emotions natural kinds?, Perspectives on Psychological
  Science 1~(1) (2006) 28--58.

\bibitem{russell1999core}
J.~A. Russell, L.~F. Barrett, Core affect, prototypical emotional episodes, and
  other things called emotion: {D}issecting the elephant., Journal of
  Personality and Social Psychology 76~(5) (1999) 805.

\bibitem{ekman2011basic}
P.~Ekman, D.~Cordaro, What is meant by calling emotions basic, Emotion Review
  3~(4) (2011) 364--370.

\bibitem{johnson1989language}
P.~N. Johnson-Laird, K.~Oatley, The language of emotions: {A}n analysis of a
  semantic field, Cognition and Emotion 3~(2) (1989) 81--123.

\bibitem{rosch1975cognitive}
E.~Rosch, Cognitive representations of semantic categories., Journal of
  Experimental Psychology: General 104~(3) (1975) 192.

\bibitem{boster1988natural}
J.~S. Boster, Natural sources of internal category structure: Typicality,
  familiarity, and similarity of birds, Memory \& Cognition 16~(3) (1988)
  258--270.

\bibitem{garrard2001prototypicality}
P.~Garrard, M.~A. Lambon~Ralph, J.~R. Hodges, K.~Patterson, Prototypicality,
  distinctiveness, and intercorrelation: {A}nalyses of the semantic attributes
  of living and nonliving concepts, Cognitive Neuropsychology 18~(2) (2001)
  125--174.

\bibitem{lieberman2007quantifying}
E.~Lieberman, J.-B. Michel, J.~Jackson, T.~Tang, M.~A. Nowak, Quantifying the
  evolutionary dynamics of language, Nature 449~(7163) (2007) 713--716.

\bibitem{pagel2007frequency}
M.~Pagel, Q.~D. Atkinson, A.~Meade, Frequency of word-use predicts rates of
  lexical evolution throughout {I}ndo-{E}uropean history, Nature 449~(7163)
  (2007) 717--720.

\bibitem{pagel2013ultraconserved}
M.~Pagel, Q.~D. Atkinson, A.~S. Calude, A.~Meade, Ultraconserved words point to
  deep language ancestry across {E}urasia, Proceedings of the National Academy
  of Sciences 110~(21) (2013) 8471--8476.

\bibitem{vejdemo2016semantic}
S.~Vejdemo, T.~H{\"o}rberg, Semantic factors predict the rate of lexical
  replacement of content words, PLOS ONE 11~(1) (2016) e0147924.

\bibitem{monaghan2019cognitive}
P.~Monaghan, S.~G. Roberts, Cognitive influences in language evolution:
  Psycholinguistic predictors of loan word borrowing, Cognition 186 (2019)
  147--158.

\bibitem{hamilton2016diachronic}
W.~L. Hamilton, J.~Leskovec, D.~Jurafsky, Diachronic word embeddings reveal
  statistical laws of semantic change, in: Proceedings of the 54th Annual
  Meeting of the Association for Computational Linguistics, 2016.

\bibitem{dubossarsky2017outta}
H.~Dubossarsky, D.~Weinshall, E.~Grossman, Outta control: Laws of semantic
  change and inherent biases in word representation models, in: Proceedings of
  the 2017 Conference on Empirical Methods in Natural Language Processing,
  2017, pp. 1136--1145.

\bibitem{boyd1988culture}
R.~Boyd, P.~J. Richerson, Culture and the evolutionary process, University of
  Chicago press, 1988.

\bibitem{bybee2007frequency}
J.~Bybee, et~al., Frequency of use and the organization of language, Oxford
  University Press on Demand, 2007.

\bibitem{geeraerts1997diachronic}
D.~Geeraerts, Diachronic Prototype Semantics: A Contribution to Historical
  Lexicology, Oxford University Press, 1997.

\bibitem{ramiro2018algorithms}
C.~Ramiro, M.~Srinivasan, B.~C. Malt, Y.~Xu, Algorithms in the historical
  emergence of word senses, Proceedings of the National Academy of Sciences
  115~(10) (2018) 2323--2328.

\bibitem{dubossarsky2015bottom}
H.~Dubossarsky, Y.~Tsvetkov, C.~Dyer, E.~Grossman, A bottom up approach to
  category mapping and meaning change., in: NetWordS, 2015, pp. 66--70.

\bibitem{mikolovA}
T.~Mikolov, K.~Chen, G.~S. Corrado, J.~Dean, Efficient estimation of word
  representations in vector space, in: Workshop Papers in {I}nternational
  {C}onference on {M}achine {L}earning, 2013.

\bibitem{mikolov2013distributed}
T.~Mikolov, I.~Sutskever, K.~Chen, G.~S. Corrado, J.~Dean, Distributed
  representations of words and phrases and their compositionality, in: Advances
  in Neural Information Processing Systems, 2013, pp. 3111--3119.

\bibitem{caliskan2017semantics}
A.~Caliskan, J.~J. Bryson, A.~Narayanan, Semantics derived automatically from
  language corpora contain human-like biases, Science 356~(6334) (2017)
  183--186.

\bibitem{kulkarni2015statistically}
V.~Kulkarni, R.~Al-Rfou, B.~Perozzi, S.~Skiena, Statistically significant
  detection of linguistic change, in: Proceedings of the 24th International
  Conference on World Wide Web, 2015.

\bibitem{xie19}
J.~Xie, R.~Ferreira Pinto~Jr., G.~Hirst, Y.~Xu, Text-based inference of moral
  sentiment change, in: Proceedings of the 2019 Conference on Empirical Methods
  in Natural Language Processing, 2019.

\bibitem{li2019macroscope}
Y.~Li, T.~Engelthaler, C.~S. Siew, T.~T. Hills, The macroscope: A tool for
  examining the historical structure of language, Behavior Research Methods
  51~(4) (2019) 1864--1877.

\bibitem{buechel2018word}
S.~Buechel, U.~Hahn, Word emotion induction for multiple languages as a deep
  multi-task learning problem, in: Proceedings of the 2018 Conference of the
  North American Chapter of the Association for Computational Linguistics:
  Human Language Technologies, 2018, pp. 1907--1918.

\bibitem{mohammad2016sentiment}
S.~M. Mohammad, Sentiment analysis: {D}etecting valence, emotions, and other
  affectual states from text, in: Emotion Measurement, 2016, pp. 201--237.

\bibitem{calvo2013emotions}
R.~A. Calvo, S.~Mac~Kim, Emotions in text: dimensional and categorical models,
  Computational Intelligence 29~(3) (2013) 527--543.

\bibitem{xu2015computational}
Y.~Xu, C.~Kemp, A computational evaluation of two laws of semantic change., in:
  Proceedings of the 37th Annual Meeting of the Cognitive Science Society,
  2015.

\bibitem{niedenthal2004prototype}
P.~Niedenthal, C.~Auxiette, A.~Nugier, N.~Dalle, P.~Bonin, M.~Fayol, A
  prototype analysis of the french category “{\'e}motion”, Cognition and
  Emotion 18~(3) (2004) 289--312.

\bibitem{reed1972pattern}
S.~K. Reed, Pattern recognition and categorization, Cognitive Psychology 3~(3)
  (1972) 382--407.

\bibitem{ashby1995categorization}
F.~G. Ashby, L.~A. Alfonso-Reese, Categorization as probability density
  estimation, Journal of Mathematical Psychology 39~(2) (1995) 216--233.

\bibitem{seabold2010statsmodels}
S.~Seabold, J.~Perktold, statsmodels: Econometric and statistical modeling with
  python, in: 9th Python in Science Conference, 2010.

\bibitem{wold1987principal}
S.~Wold, K.~Esbensen, P.~Geladi, Principal component analysis, Chemometrics and
  Intelligent Laboratory Systems 2~(1-3) (1987) 37--52.

\bibitem{rosch1978cognition}
E.~Rosch, Principles of categorization, in: E.~Rosch, B.~B. Lloyd (Eds.),
  Cognition and Categorization, Lawrence Erlbaum Associates, Hillsdale, NJ,
  1978, pp. 27--48.

\bibitem{luo2018stability}
Y.~Luo, Y.~Xu, Stability in the temporal dynamics of word meanings., in:
  Proceedings of the 40th Annual Meeting of the Cognitive Science Society,
  2018.

\bibitem{xu2019wordforms}
A.~Xu, C.~Ramiro, Y.~Xu, A predictability-distinctiveness trade-off in the
  historical emergence of word forms., in: Proceedings of the 41st Annual
  Meeting of the Cognitive Science Society, 2019.

\bibitem{HTE4.2}
C.~Kay, J.~Roberts, M.~Samuels, I.~Wotherspoon,
  \href{http://historicalthesaurus.arts.gla.ac.uk/}{The {H}istorical
  {T}hesaurus of {E}nglish, version 4.21} (2017).
\newline\urlprefix\url{http://historicalthesaurus.arts.gla.ac.uk/}

\bibitem{monaghan2014age}
P.~Monaghan, Age of acquisition predicts rate of lexical evolution, Cognition
  133~(3) (2014) 530--534.

\bibitem{fellbaum1998}
C.~Fellbaum, WordNet: An Electronic Lexical Database, MIT Press, Cambridge, MA,
  1998.

\bibitem{nltk}
E.~Loper, S.~Bird, Nltk: The natural language toolkit, in: In Proceedings of
  the ACL Workshop on Effective Tools and Methodologies for Teaching Natural
  Language Processing and Computational Linguistics. Philadelphia: Association
  for Computational Linguistics, 2002.

\bibitem{dale1981living}
E.~Dale, J.~O'rourke, The Living Word Vocabulary, the Words We Know: A National
  Vocabulary Inventory., Chicago: World Book, 1981.

\bibitem{brysbaert2017test}
M.~Brysbaert, A.~Biemiller, Test-based age-of-acquisition norms for 44 thousand
  english word meanings, Behavior research methods 49~(4) (2017) 1520--1523.

\end{thebibliography}

\end{document}